\documentclass[conference]{IEEEtran}
\usepackage{cite}
\usepackage{amsmath,amssymb,amsfonts}
\usepackage{algorithmic}
\usepackage{graphicx}
\usepackage{textcomp}
\usepackage{booktabs}
\usepackage{hyperref}
\usepackage[dvipsnames]{xcolor}
\begin{document}

\title{BanglaEmbed: Efficient Sentence Embedding Models for a Low-Resource Language Using Cross-Lingual Distillation Techniques}

\author{\IEEEauthorblockN{Muhammad Rafsan Kabir}
\IEEEauthorblockA{\textit{Electrical and Computer Engineering} \\
\textit{North South University}\\
Dhaka, Bangladesh \\
muhammad.kabir@northsouth.edu}
\and
\IEEEauthorblockN{Md. Mohibur Rahman Nabil}
\IEEEauthorblockA{\textit{Electrical and Computer Engineering} \\
\textit{North South University}\\
Dhaka, Bangladesh \\
mohibur.nabil@northsouth.edu}
\and
\IEEEauthorblockN{Mohammad Ashrafuzzaman Khan}
\IEEEauthorblockA{\textit{Electrical and Computer Engineering} \\
\textit{North South University}\\
Dhaka, Bangladesh \\
mohammad.khan02@northsouth.edu}

}

\maketitle

\begin{abstract}
Sentence-level embedding is essential for various tasks that require understanding natural language. Many studies have explored such embeddings for high-resource languages like English. However, low-resource languages like Bengali (a language spoken by almost two hundred and thirty million people) are still under-explored. This work introduces two lightweight sentence transformers for the Bangla language, leveraging a novel cross-lingual knowledge distillation approach. This method distills knowledge from a pre-trained, high-performing English sentence transformer. Proposed models are evaluated across multiple downstream tasks, including paraphrase detection, semantic textual similarity (STS), and Bangla hate speech detection. The new method consistently outperformed existing Bangla sentence transformers. Moreover, the lightweight architecture and shorter inference time make the models highly suitable for deployment in resource-constrained environments, making them valuable for practical NLP applications in low-resource languages.
\end{abstract}


\section{Introduction}
In recent years, there have been remarkable advancements in the field of natural language processing (NLP) \cite{khurana2023natural}. One significant development is the emergence of high-quality sentence embedding models \cite{reimers-gurevych-2019-sentence}, which can effectively map sentences into an embedding space based on their meaning and context. However, low-resource languages like Bangla still lag behind high-resource languages like English in terms of the quality of sentence transformers. This disparity exists primarily due to the limited availability of high-quality data in languages like Bangla.

Low-resource languages lack high-quality, large text corpora necessary for developing effective sentence embedding models \cite{king2015practical}. To address this gap, our work proposes an approach that utilizes a machine translation dataset \cite{wang2022progress} instead of a large text corpus to train lightweight sentence transformers for the Bangla language. 
In this study, we introduce two distinct embedding models \cite{reimers-gurevych-2019-sentence} for Bangla, trained using an English-Bangla machine translation dataset \cite{hasan-etal-2020-low} and two different loss functions—mean squared error (MSE) and multiple negatives ranking loss \cite{henderson2017efficient}.

\begin{figure}[!t]
    \centering
    \includegraphics[width=\linewidth]{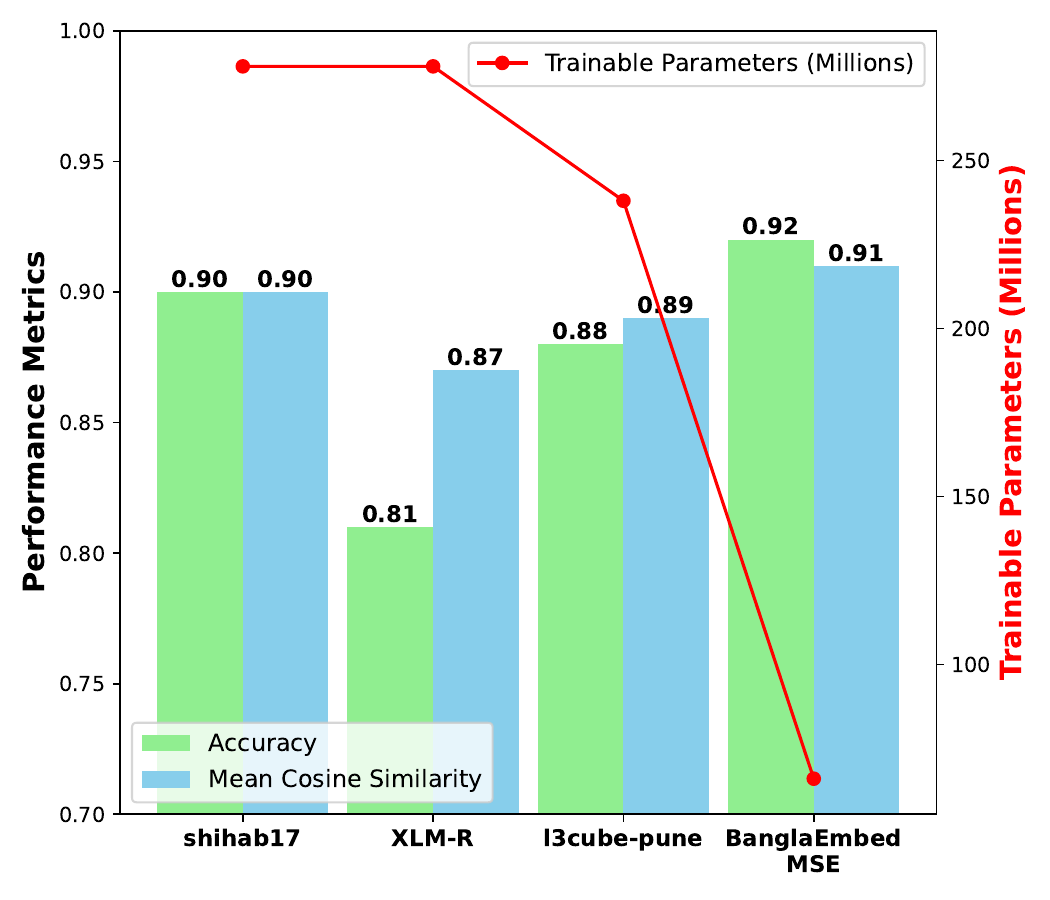}
    \caption{Performance comparison of our proposed sentence transformer, \texttt{BanglaEmbed-MSE}, on the paraphrase detection task, evaluated using accuracy, mean cosine similarity, and the number of trainable parameters.}
    \label{fig1}
\end{figure}

In our study, we employed a knowledge distillation pipeline \cite{hinton2015distilling} that utilizes a pre-trained teacher sentence embedding model alongside a custom student sentence embedding model. Since Bangla lacks a high-performing pre-trained sentence embedding model, we used an English pre-trained sentence transformer as the teacher model. As an English sentence and its translated Bangla counterpart convey the same contextual meaning, the embeddings for both sentences should map to the same embedding space. This approach allows us to supervise the training of lightweight Bangla sentence transformers (student models) using a pre-trained English sentence transformer (teacher model). Our proposed methodology facilitates the training of an embedding model for a low-resource language like Bangla by leveraging a pre-trained English sentence transformer and a machine translation dataset, rather than relying on a large text corpus. Specifically, we used the English-Bangla Machine Translation dataset introduced by \cite{hasan-etal-2020-low} for training. Figure \ref{fig1} presents a performance comparison of our introduced sentence embedding model, \texttt{BanglaEmbed-MSE}, with other existing Bangla sentence transformers on the paraphrase detection task. 
The significant contributions of this work are as follows: 

\begin{itemize}
    \item Introduction of two lightweight sentence transformers for a low-resource language, Bangla, utilizing mean squared error (MSE) and multiple negatives ranking loss. 
    \item The novel training approach employs cross-lingual knowledge distillation and leverages a pre-trained model from a high-resource language to overcome the lack of large Bangla corpora, using a smaller translation dataset.
    \item The proposed models are evaluated against existing Bangla sentence transformers across multiple downstream tasks, demonstrating superior performance with reduced computational requirements.
\end{itemize}

\section{Related Works}
\noindent \textbf{Sentence Embeddings.}
Sentence embedding plays a crucial role in modern language models, capturing semantic meaning that informs the reasoning abilities of these models. Numerous efforts have been made to improve sentence embeddings, particularly in high-resource languages like English \cite{conneau-etal-2017-supervised, cer2018universal}. However, significant gaps remain for low-resource languages. To address this, various multilingual approaches \cite{reimers-gurevych-2020-making, saraswat2021text}, have been proposed to extend sentence representation learning to a wider set of languages. \\

\noindent \textbf{Multilingual Sentence Embeddings.}
The Multilingual Universal Sentence Encoder (mUSE) \cite{chidambaram2018learning} employs a dual encoder architecture and was trained using a multi-task learning approach on the SNLI dataset \cite{bowman-etal-2015-large}. The model's training included a translation ranking task, where it had to identify the correct translation from a set of candidates. To improve performance, it used hard negative samples—sentences similar to the correct translation but not entirely accurate. Trained on 16 languages, the model showed strong effectiveness in multilingual tasks. For low-resource language translation tasks, Gao et al. \cite{gao-etal-2021-simcse} employed the pre-trained SimCSE model to generate sentence embeddings, which were combined with representations using an embed-fusion module. This output was fed into the encoder-decoder attention layer of the model's decoder block. During training, only the parameters of the embed-fusion module and transformer model were updated, while the SimCSE parameters were frozen. During inference, the SimCSE and embed-fusion modules acted as plug-ins, ensuring compatibility with various model architectures. \\
 
\noindent \textbf{Distillation for Sentence Transformers.}
Knowledge distillation (KD) has proven to be an effective approach for transferring knowledge from a large, high-capacity model (teacher) to a smaller, more efficient model (student) \cite{hinton2015distilling}. In \cite{Xie2023}, SimTDE was introduced, employing knowledge distillation to train a student model with a shallower token embedding block and fewer transformer layers than the teacher model. Despite its simpler architecture, the student model achieved performance comparable to the teacher while offering twice the inference efficiency. Similarly, Reimers and Gurevych \cite{reimers-gurevych-2020-making} used a KD-based approach, where a pre-trained English sentence transformer served as the teacher model, and a lightweight student model was trained to learn sentence representations for both English and additional languages. This process enabled the student model to capture cross-lingual representations under the guidance of the more robust teacher model.

\section{Methodology}
\subsection{Problem Formulation}
Let \( D \) be a machine-translation dataset, where each data point consists of a Bangla sentence $BN_i$ and its corresponding English translation $EN_i$, i.e., \( D = \{BN_i, EN_i\}_{i=1}^N \). When an English sentence $EN_i$ is passed through the teacher model $T$, it generates an embedding $E_T(EN_i)$, which can be considered the \emph{target embedding}, denoted as \( \mathbf{y}_i \). Similarly, when the corresponding Bangla sentence \( BN_i \) is passed through the student model \( S \), the model produces an embedding \( E_S(BN_i) \), which can be considered the \emph{predicted embedding}, denoted as \( \hat{\mathbf{y}}_i \).
The goal is to optimize the parameters of the student model \( S \) such that the distance between the \emph{predicted embedding} \( \hat{\mathbf{y}}_i = E_S(BN_i) \) and the \emph{target embedding} \( \mathbf{y}_i = E_T(EN_i) \) is minimized. 

\begin{equation}
L = \frac{1}{N} \sum_{i=1}^{N} \text{distance}\left( \mathbf{y}_i, \hat{\mathbf{y}}_i \right)
\end{equation}

Here, \emph{distance} represents a metric that quantifies the difference between the teacher’s and student’s embeddings. By minimizing this distance, the student model \( S \) learns to align its output embeddings \( \hat{\mathbf{y}}_i = E_S(BN_i) \) as closely as possible to the teacher's embeddings \( \mathbf{y}_i = E_T(EN_i) \).

\subsection{Training Pipeline for \texttt{BanglaEmbed}}
Low-resource languages like Bangla still face a shortage of high-performing sentence embedding models compared to high-resource languages. In this study, we introduce two distinct Bangla sentence transformers: \texttt{BanglaEmbed-MNR} and \texttt{BanglaEmbed-MSE}. These sentence transformers are trained using a cross-lingual knowledge distillation approach on an English-Bangla machine translation dataset. \\

\begin{figure}[!t]
    \centering
    \includegraphics[width=\linewidth]{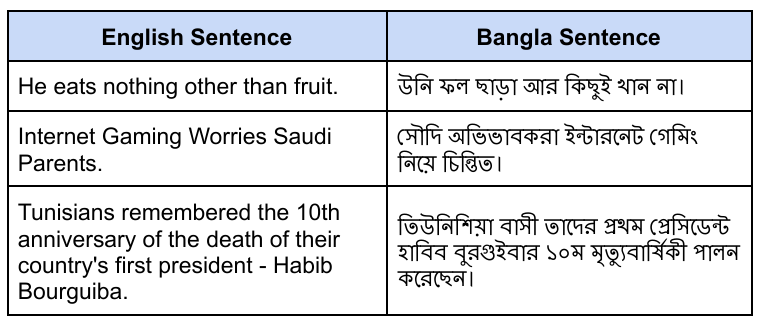}
    \caption{Sample EN-BN sentence pairs from the machine translation dataset.}
    \label{sample_data}
\end{figure}

\noindent \textbf{Training Data.}
\label{data}
For training purposes, we utilized the publicly available BanglaNMT machine translation dataset \cite{hasan-etal-2020-low}, which consists of parallel English and Bangla statements. The initial size of the corpus was 2.75 million sentence pairs. After pre-processing, the dataset was reduced to approximately 2.66 million sentence pairs, which is sufficient for training a sentence transformer. This dataset is the largest available machine translation resource for the Bangla language. The original and the pre-processed datasets are accessible through the official GitHub repository\footnote{\url{https://github.com/csebuetnlp/banglanmt}}. A few sample pairs from the training dataset are shown in Figure \ref{sample_data}. \\

\noindent \textbf{Knowledge Distillation.}
Instead of following the standard training approach, we adopted a unique strategy for training sentence transformers in a low-resource language by leveraging knowledge from a pre-trained sentence transformer in a high-resource language, as proposed by \cite{reimers-gurevych-2020-making}. To implement this, we utilized a knowledge distillation pipeline \cite{hinton2015distilling}, creating a teacher-student framework between a high-performing English sentence transformer (teacher) and a custom lightweight Bangla sentence transformer (student).

\begin{figure*}[!t]
    \centering
    \includegraphics[width=0.8\linewidth]{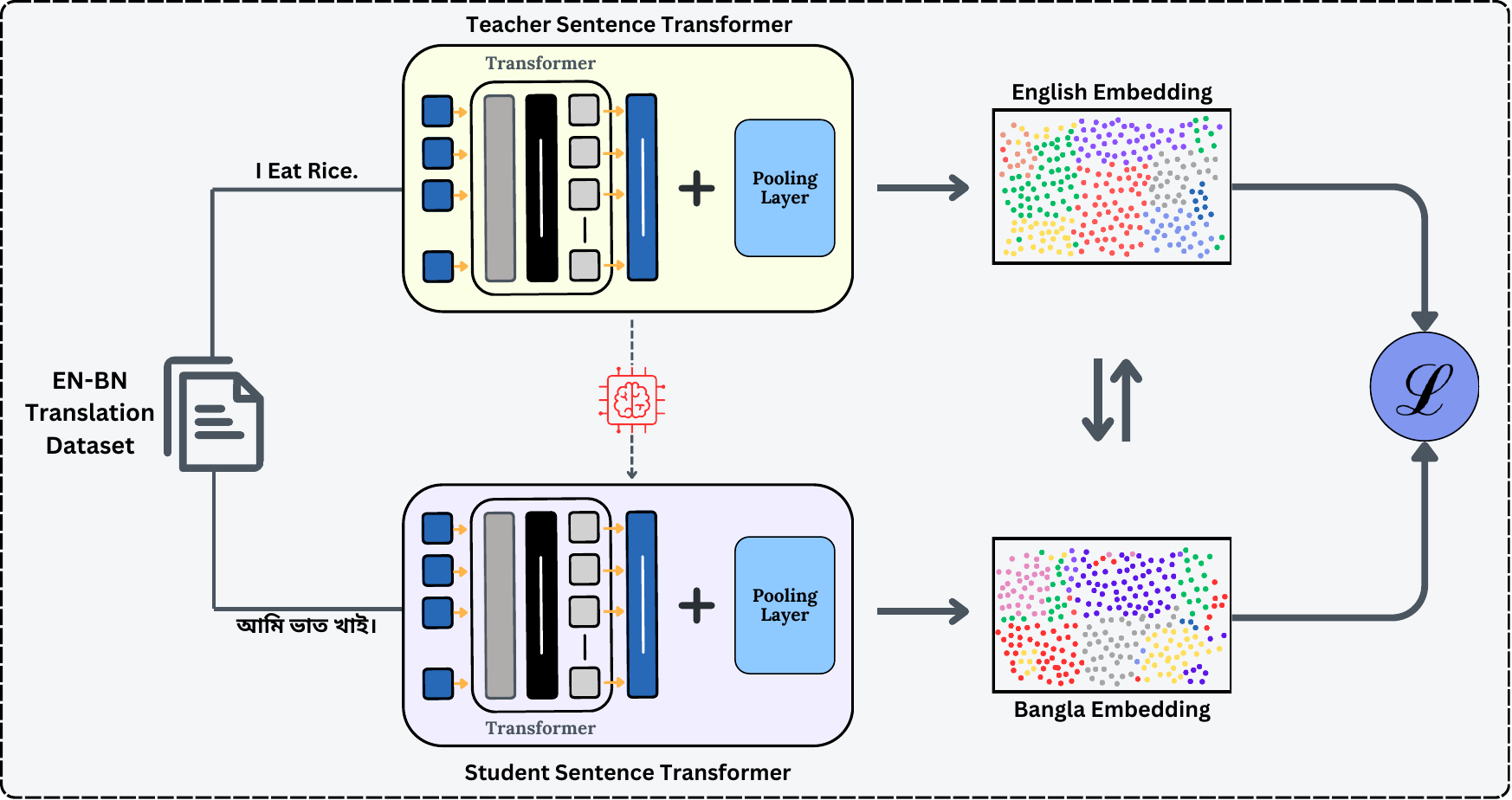}
    \caption{Proposed cross-lingual knowledge distillation methodology for training the Bangla sentence transformer, leveraging an English-Bangla machine translation dataset. The bidirectional arrow ($\downarrow \uparrow$) indicates that both English and Bangla embeddings are aligned to map into the same embedding space.}
    \label{methodology}
\end{figure*}

In this process, the pre-trained English sentence transformer (teacher) was provided with English statements from the machine translation dataset (discussed in Section \ref{data}), while the custom lightweight embedding model (student) was fed the corresponding Bangla statements. Since the English and Bangla embeddings represent the same contextual meaning, they should ideally map to the same embedding space. The student model's output embeddings were supervised using the embeddings generated by the pre-trained teacher model. Consequently, knowledge was distilled from the pre-trained English sentence transformer into the lightweight Bangla sentence transformer, completing the distillation process. This cross-lingual distillation method allows for the training of a Bangla sentence embedding model without requiring an extremely large text corpus. Figure \ref{methodology} illustrates the training pipeline used in this study. \\

\noindent \textbf{Loss Functions.}
We employed two different loss functions during the training phase. Specifically, we utilized Mean Squared Error (MSE) and Multiple Negatives Ranking Loss for training \texttt{BanglaEmbed-MSE} and \texttt{BanglaEmbed-MNR}, respectively.

\begin{enumerate}
    \item \textbf{Mean Squared Error:} To train the \texttt{BanglaEmbed-MSE} sentence transformer, the MSE loss function was used, which calculates the squared difference between the embeddings from the English transformer (teacher) and the Bangla transformer (student), as shown in equation \ref{mse_loss}. The goal is to minimize this loss by aligning the two embeddings in the embedding space.

\begin{equation}
\label{mse_loss}
    L_{MSE} = \frac{1}{N} \sum_{n=1}^{N} (E_{T}^{i} - E_{S}^{i})^2
\end{equation}

Here, $E_T$ and $E_S$ refer to the embeddings generated by the teacher and student models, respectively, and $N$ represents the total number of sentence pairs. \\

\item \textbf{Multiple Negatives Ranking Loss:} This loss function is employed to train \texttt{BanglaEmbed-MNR}. The objective of this ranking loss is to minimize the distance between the teacher embedding and the positive sample (correct translation), while maximizing the distance between the teacher embedding and the negative samples (incorrect translations) in the embedding space. During training, one sample from the batch is considered the positive sample, while the remaining samples are treated as negative examples. Equation \ref{mnr_loss} illustrates the multiple negatives ranking loss.
\begin{equation}
\label{mnr_loss}
    L_{MNR} = -\log \frac{e^{(\cos(E_T, E_{S}^{+}))}}{e^{(\cos(E_T, E_{S}^{+}))} + \sum_{n=1}^{K} e^{(\cos(E_T, E_{S}^{-(n)}))}}
\end{equation}

Here, $E_{S}^{+}$ refers to the positive sample, and $E_{S}^{-(n)}$ refers to the $n$-th negative sample, and $K$ represents the total number of negative samples.
\end{enumerate}

\subsection{Model Overview}

For our training pipeline, we utilize two models—a teacher and a student. The teacher model is a pre-trained sentence transformer capable of generating high-quality embeddings for English sentences. Specifically, we use the pre-trained \textbf{multi-qa-distilbert-cos-v1} from SBERT, which contains 66.4 million parameters. This model has been trained on 215 million diverse question-answer pairs, making it highly effective for generating robust sentence embeddings.

The student models, \texttt{BanglaEmbed-MSE} and \texttt{BanglaEmbed-MNR}, are custom embedding models based on the \textbf{distilbert-base-uncased} architecture \cite{Sanh2019DistilBERTAD}, a lightweight variant of BERT \cite{devlin-etal-2019-bert}. Like the teacher model, the student model also has 66.4 million parameters. The student model was adapted for sentence embedding tasks by adding a pooling layer that averages token embeddings, resulting in compact and efficient sentence representations. The lightweight student model, along with its pooling mechanism, makes it ideal for downstream tasks in low-resource languages like Bangla.

\section{Experiments}
\subsection{Setup}

\noindent \textbf{Dataset.}
To evaluate the performance of the introduced sentence transformers, we assessed them on three distinct tasks: paraphrase detection, semantic textual similarity (STS), and hate-speech classification.
For the paraphrase detection task, we employed the test set from the public \emph{BanglaParaphrase} dataset by \cite{akil-etal-2022-banglaparaphrase}, which contains 23,300 sentence pairs (source and target).
For the STS evaluation, we utilized the \emph{SemEval Task 1: Semantic Textual Similarity (STS)} dataset, consisting of 250 pairs of statements with labels ranging from 0 to 5, a widely recognized gold-standard dataset \cite{cer-etal-2017-semeval}. Given the dataset's high quality, we translated it into Bangla rather than using a lower-quality Bangla STS dataset. The English sentence pairs were translated into Bangla using GPT-4o, followed by manual human validation, while preserving the original labels.
Finally, we conducted a qualitative evaluation on the Bangla hate-speech classification task using the \emph{Bengali Hate Speech Dataset} introduced by \cite{karim2020BengaliNLP}. This dataset contains 3,420 instances and categorizes hate speech into five distinct types: personal, political, geopolitical, gender abuse, and religious hate speech. \\

\noindent \textbf{Implementation Details.}
We introduced two versions of a sentence transformer for Bangla, a low-resource language. These two versions were trained on the BanglaNMT machine translation dataset \cite{hasan-etal-2020-low}, using mean squared error and multiple negatives ranking loss, respectively. Both sentence transformers were trained for 10 epochs with a batch size of 4, utilizing the \emph{SentenceTransformerTrainer} from Sentence Transformers (SBERT) for efficient model training. The experiments were implemented using the \emph{PyTorch} framework. All training and evaluation were conducted on an NVIDIA RTX 3080 GPU with 12 GB of VRAM. Table \ref{hyperparameters} presents the hyperparameters employed during our experiments. \\

\begin{table}[!t]
\caption{Hyperparameters used during our experiments.}
    \label{hyperparameters}
    \centering
\setlength{\tabcolsep}{2.2em}
    \begin{tabular}{l l}
    \toprule
        \textbf{Hyperparameter} & \textbf{Value} \\
        \midrule
         Epochs & 10 \\
         Batch Size & 4 \\
         Warmup Ratio & 0.1 \\
         Learning Rate & 5e-5 \\
         Optimizer & AdamW \\
         \bottomrule
    \end{tabular}
\end{table}

\noindent \textbf{Evaluation.}
We used three strategies to evaluate the performance of our sentence transformers, alongside other available Bangla sentence transformers. First, we evaluated all models on the Bangla paraphrase detection task by calculating the Mean Cosine Similarity (MCS) between source and target embeddings. Classification accuracy was also measured using a cosine similarity threshold of 0.8; pairs with cosine similarity $\geq$ 0.8 were classified as paraphrases, while those below were classified as non-paraphrases. 
Second, we assessed the sentence transformers' performance on the Semantic Textual Similarity (STS) task, where the objective was to assign a similarity score (ranging from 0 to 5) to pairs of Bangla sentences. We used two evaluation metrics for this task: Pearson correlation ($r$) and Spearman correlation ($\rho$).
Lastly, we conducted a qualitative evaluation by generating t-SNE clustering plots \cite{van2008visualizing} for a Bangla hate-speech classification task consisting of five distinct classes. By examining the separation and clustering of the generated t-SNE plots, we compared the quality of the embeddings given by each of the sentence transformers.

\begin{table*}[ht]
\small
\centering
\caption{Comparison of evaluation results for two downstream tasks: Paraphrase Detection and Semantic Textual Similarity (STS). Here, \emph{MCS} denotes Mean Cosine Similarity.}\label{results}

\renewcommand{\arraystretch}{1.3}
\setlength{\tabcolsep}{0.8em}
\begin{tabular}{l c c c c c c}
\toprule
\textbf{Models} & \textbf{Parameters} & \textbf{Inference Time} & \multicolumn{2}{c}{\textbf{Paraphrase Detection}} & \multicolumn{2}{c}{\textbf{STS}} \\
\cmidrule(rl){4-5} \cmidrule(rl){6-7}
& & & MCS & Accuracy & $\rho$ & $r$\\

\midrule

Bangla Sentence Transformer \footnotemark & 278 M & 13.4 s & 0.90 & 0.90 & 0.65 & 0.65 \\
XLM-R 100L BERT NLI STSB \cite{reimers-gurevych-2019-sentence}& 278 M & 13.9 s & 0.87 & 0.81 & 0.65 & 0.64 \\ 
BengaliSBERT-STS \cite{deode2023l3cube} & 238 M & 12.7 s & 0.89 & 0.88 & 0.72 & \textbf{0.72}\\
BanglaEmbed-MNR (\textcolor{BlueViolet}{Ours}) & 66 M & 12.3 s & 0.87 & 0.82 & 0.63 & 0.62 \\
BanglaEmbed-MSE (\textcolor{BlueViolet}{Ours}) & 66 M & \textbf{12.0 s} & \textbf{0.91} & \textbf{0.92} & \textbf{0.73} & 0.70 \\ 

\bottomrule
\end{tabular}
\end{table*}

\subsection{Results and Analysis}
\begin{figure*}[!t]
    \centering
    \includegraphics[width=0.79\linewidth]{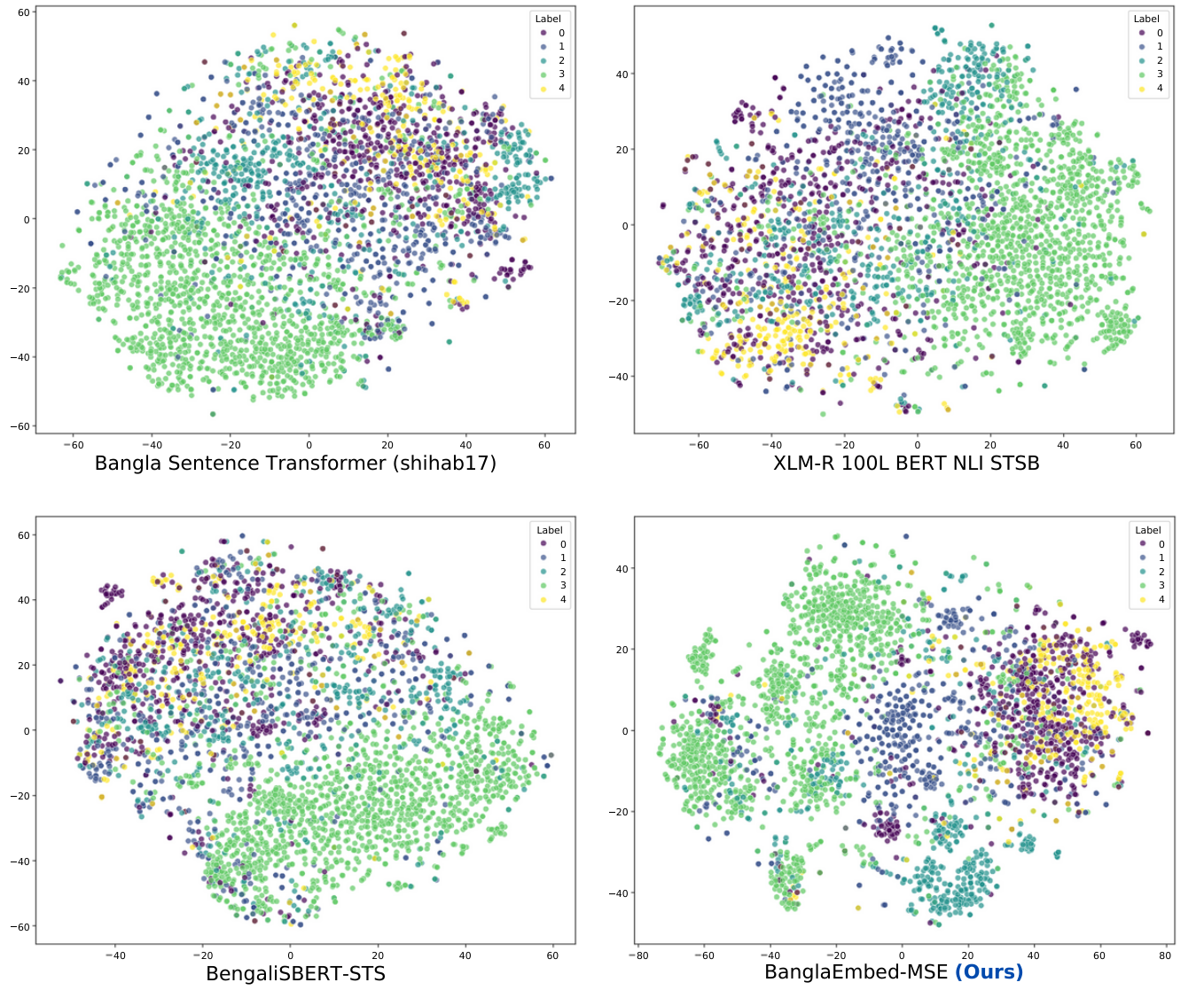}
    \caption{t-SNE visualizations of four distinct sentence transformers. The \texttt{BanglaEmbed-MSE} model shows superior performance in separating clusters, indicating higher quality sentence embeddings compared to the other models.}
    \label{tsne}
\end{figure*}

We compare the performance of our proposed Bangla embedding models across three downstream tasks: Paraphrase Detection, Semantic Textual Similarity (STS), and Hate Speech Classification. Table \ref{results} presents the results for the paraphrase detection and STS tasks, achieved by both the introduced sentence embedding models and existing embeddings for the Bangla language. Additionally, the table includes the inference time required to compute embeddings for the source sentences in the BanglaParaphrase dataset \cite{akil-etal-2022-banglaparaphrase}.

\footnotetext{\url{https://huggingface.co/shihab17/bangla-sentence-transformer}}

From Table \ref{results}, we observe the following: \textbf{\emph{(1)}} Both of the introduced embedding models, \texttt{BanglaEmbed-MNR} and \texttt{BanglaEmbed-MSE}, are lightweight, each with approximately 66 million parameters, making them computationally efficient. This reduced parameter count also results in shorter inference times compared to other existing models.
\textbf{\emph{(2)}} \texttt{BanglaEmbed-MSE} achieves the best performance in paraphrase detection, with a mean cosine similarity of 0.91 and an accuracy of 0.92, outperforming all other models. \textbf{\emph{(3)}} In the semantic textual similarity (STS) task, \texttt{BanglaEmbed-MSE} shows strong performance with a Spearman correlation of 0.73 and Pearson correlation of 0.70, while remaining efficient with fewer parameters. \textbf{\emph{(4)}} Although \texttt{BanglaEmbed-MNR} does not surpass the highest-performing models, it remains competitive, offering computational efficiency with fewer parameters and a shorter inference time.

For the hate speech classification task, we conducted a qualitative evaluation using t-SNE visualizations, as shown in Figure \ref{tsne}. This visualization compares the clustering produced by four different sentence embeddings. As illustrated, \texttt{BanglaEmbed-MSE} shows superior clustering and clearer separation of hate speech classes, indicating the generation of higher-quality embeddings compared to the other models.

\section{Conclusion}
In this paper, we introduced two novel Bangla sentence transformers, \texttt{BanglaEmbed-MSE} and \texttt{BanglaEmbed-MNR}, which outperform existing models with fewer parameters. By employing cross-lingual knowledge distillation, we successfully trained these models for the Bangla language, making a significant contribution to advancing natural language understanding in this low-resource language. While our models demonstrate superior performance on multiple downstream tasks, they also establish a strong baseline for future work in Bangla NLP. However, there remains room for improvement. Future efforts could focus on exploring more efficient architectures and training on more diverse and larger datasets to further enhance the models’ performance.

\bibliographystyle{IEEEtran}
\bibliography{example}

\end{document}